\documentclass[10pt]{article} 

\usepackage[preprint]{rlc}

\usepackage{amssymb}            
\usepackage{mathtools}          
\usepackage{mathrsfs}           
\usepackage{graphicx}           
\usepackage{subcaption}         
\usepackage[space]{grffile}     
\usepackage{url}                

\usepackage{algorithm}
\usepackage{algpseudocode}
\usepackage{amsmath}

\title{Towards Learning Foundation Models for Heuristic Functions to Solve Pathfinding Problems}

\author{Vedant Khandelwal  \\
    vedant@email.sc.edu \\
    AI Institute\\
    University of South Carolina
    \And
    Amit Sheth \\
    amit@sc.edu\\
    AI Institute \\
    University of South Carolina
    \And
    Forest Agostinelli \\
    foresta@cse.sc.edu\\
    AI Institute \\
    University of South Carolina}

\begin{document}

\maketitle

\begin{abstract}
Pathfinding problems are found throughout robotics, computational science, and natural sciences. Traditional methods to solve these require training deep neural networks (DNNs) for each new problem domain, consuming substantial time and resources. This study introduces a novel foundation model, leveraging deep reinforcement learning to train heuristic functions that seamlessly adapt to new domains without further fine-tuning. Building upon DeepCubeA, we enhance the model by providing the heuristic function with the domain's state transition information, improving its adaptability. Utilizing a puzzle generator for the 15-puzzle action space variation domains, we demonstrate our model's ability to generalize and solve unseen domains. We achieve a strong correlation between learned and ground truth heuristic values across various domains, as evidenced by robust R-squared and Concordance Correlation Coefficient metrics.  These results underscore the potential of foundation models to establish new standards in efficiency and adaptability for AI-driven solutions in complex pathfinding problems.
\end{abstract}


\section{Introduction}
\label{sec:intro}
Pathfinding aims to find a sequence of actions that forms a path from a given start state to a given goal, where a goal is a set of states considered goal states while attempting to minimize the path's cost. Pathfinding problems are found throughout computer science, robotics, mathematics, and the natural sciences. Heuristic search is one of the most prominent class of solutions to pathfinding problems. A key component of heuristic search is the heuristic function, which maps states to estimates of the cost of the shortest path from the given state to the nearest goal state, also known as the ``cost-to-go''. Recently, deep reinforcement learning \citep{sutton2018reinforcement} is a promising method for the automatic construction of domain-specific heuristic functions in a largely domain-independent fashion \citep{agostinelli2019solving,agostinelli2024specifying}. However, the time it takes to train a deep neural network (DNN) \citep{schmidhuber2015deep} heuristic function with deep reinforcement learning can take days when using multiple high-end graphics processing units (GPUs). As a result, small changes in a domain can require significant time to re-train the DNN. Furthermore, the hardware requirements limit the accessibility of these methods to the broader research community.

The aforementioned problems have also been encountered in computer vision and natural language processing fields. To address these problems, foundation models based on DNNs have been successfully constructed and employed in research and industry. Foundation models are trained on large amounts of diverse data that can be adapted to new tasks with little to no fine-tuning. Successfully creating a foundation model for heuristic functions could have a similar impact in the area of heuristic search. To accomplish this, the heuristic function would have to generalize over states and pathfinding domains.


In this study, we propose a foundation model designed to generalize across different 15-puzzle action space variation domains. To generalize across different domains, we would generally need state transition function, transition cost function, and transition function. The proposed work narrows the challenge to handling subsets of domain variations where the state transition dynamics are influenced by different action spaces. To achieve this, we incorporate action space information with the state representations, thereby implicitly representing the domain variations through the available actions. By leveraging deep reinforcement learning, our model adapts seamlessly to new domains without retraining. Strong correlations between learned heuristic values and ground truth metrics, as indicated by robust R-squared and Concordance Correlation Coefficient (CCC) values, show the model's ability to generalize on unseen domains.

Our contributions are twofold:\\
\noindent 1. We introduce a novel approach integrating state transition information with state representations to enhance the generalizability of heuristic functions in pathfinding problems.\\
\noindent 2. We demonstrate the model's ability to generalize across 15-puzzle action variation domains and achieve competitive results against traditional domain-specific heuristic methods.

\section{Background and Literature Review}
This section provides an overview of the key concepts and existing research relevant to our study. 

\subsection{Foundation Model}\label{subsec:fms}

Foundation models are pre-trained deep learning models using supervised or self-supervised techniques on extensive, diverse datasets. These models are adaptable to various downstream tasks.

\noindent\textbf{Definition:}
A foundation model \( F_\theta \) is trained to minimize the loss function \( L \) over a dataset \( D \), with diverse inputs:
\[
\theta^* = \arg \min_{\theta} L(F_\theta(x), x) \quad \text{for} \quad x \in D
\]
Once trained, \( F_\theta \) can generalize across different domains and tasks with or without fine-tuning. Foundation models have significant applications in natural language processing (NLP), computer vision, and healthcare.


\subsection{Pathfinding}
A pathfinding domain can be defined by a directed weighted graph, where nodes represent states, edges represent transitions between states, edge weights represent transition costs, and a particular pathfinding problem can be defined by a domain, start state, and goal. Since many problems can have a very large state space, representing this graph in its entirety is often infeasible. Therefore, when solving pathfinding problems, we assume knowledge of the action space, $\mathcal{A}$, which defines the actions that can be taken to transition between nodes in the graph, the state transition function, $T$, that maps states and actions to next states, and the transition cost function, $c$, which maps states and actions to the transition cost. The cost of a path is the sum of transition costs.

Heuristic search is one of the most notable solution classes for pathfinding problems, and A* search \citep{hart1968formal} is the most notable heuristic search algorithm. A* search maintains a search tree, and nodes are expanded according to a priority obtained by adding the path cost to the cost-to-go computed by the heuristic function $h$. A* search terminates when a node associated with a goal state is selected for expansion.

In traditional heuristic search, the heuristic function $h$ values were stored as a lookup table for all possible states $s$. However, this lookup table representation becomes infeasible for combinatorial puzzles with large state spaces like the 15-Puzzle. Therefore, we turn to approximate value iteration. This dynamic programming algorithm iteratively improves a cost-to-go computed by the heuristic function $h$, where $h$ is represented by a parameterized function implemented by a DNN. The DNN is trained to minimize the mean squared error between its estimation of the cost-to-go of state $s$, $h(s)$, and the updated cost-to-go estimation $h'(s)$:
\begin{equation}
    h'(s) = \min_a (c(s, T(s, a)) + h(T(s, a)))
    \label{eq:orgavi}
\end{equation}
where $T(s, a)$ is the state obtained from taking action $a$ in state $s$ and $c(s, T(s, a))$ is the cost to transition from state $s$ to state $s'$ taking action $a$ $\epsilon$ $\mathcal{A}$. 

For this study, we work with the 15-puzzle, or sliding tile puzzle \citep{keith2011vintage}. The goal is to rearrange the numbered tiles from an initial state to the goal state (shown in Figure \ref{fig:npuz}), using canonical moves (up (U), down (D), left (L), right (R)) and diagonal moves (upper-left (UL), upper-right (UR), down-left (DL), down-right (DL)). More details are in the Appendix \ref{subsec:npuzz}.

\subsection{Review of DeepCubeA}

DeepCubeA \citep{agostinelli2019solving} applies deep reinforcement learning integrated with approximate value iteration (AVI) (Equation \ref{eq:orgavi}) and a weighted A* search to solve puzzles like Rubik's Cube, N-Puzzle, Sokoban, and Lights-out. DeepCubeA learns a domain-specific heuristic function in a largely domain-independent way without human guidance.

However, despite its effectiveness, DeepCubeA faces significant challenges, such as lengthy training times and the necessity to retrain the model from scratch for a slight change in the domain or its transition dynamics, increasing computational demands for larger puzzles. These challenges underscore the need for more adaptable and generalized solutions in AI pathfinding systems that can leverage learned heuristic functions across various domains without retraining from scratch.

\subsection{Generalization in Pathfinding Problems}

\citep{chen2024learning} introduced three novel graph representations for planning tasks using Graph Neural Networks (GNNs) to learn domain-independent heuristics. Their approach mitigates issues with large grounded GNNs by leveraging lifted representations and demonstrates superior generalization to larger problems compared to models like STRIPS-HGN. However, it faces scalability issues with large graph construction. Similarly, \citep{chen2023goose} proposed the GOOSE framework, using GNNs with novel grounded and lifted graph representations for classical planning. Their heuristics outperform STRIPS-HGN and hFF in various domains but require extensive training data and struggle with very large graphs. Additionally, \citep{toyer2018action} utilized GNNs to improve coverage and plan quality in classical planning tasks through new graph representations, though their approach only generalizes across a subset of test domains.
However, these approaches rely on supervised learning, which assumes the ability to solve moderately difficult problem instances with existing solvers, which may not always be the case for real-world problems.


Large language models (LLMs), pre-trained on extensive textual datasets, have shown potential in downstream natural language processing tasks. While attempts to use LLMs for pathfinding via in-context learning have shown modest performance \citep{sermanet2023robovqa, li2023human, silver2023generalized}, they face notable challenges and lack inherent search capabilities. 
More details are provided in the Appendix \ref{subsec:llms}.


The Fast Downward planner \citep{helmert2006fast} uses domain-independent heuristics like the Fast Forward (FF) heuristic \citep{hoffmann2001ff} to solve pathfinding problems efficiently. These heuristics generalize across various domains but do not perform as well as learned heuristics \citep{agostinelli2024specifying}. 
More details are provided in the Appendix \ref{subsec:domainindeependent}.

\section{Theoretical Framework}

Pathfinding problems are represented as weighted directed graphs, but encoding the entire graph is impractical due to its large size. To generalize across a domain, we need to understand several key components: the state space, the state transition function, and the transition cost function.

 In this study, we focus on a subset of this broader problem to demonstrate that deep reinforcement learning can achieve this. Specifically, we keep the state representation consistent, assume prior knowledge of the action space, and use a uniform transition cost function. Consequently, the state transition function is implicitly defined by the action space.

For the 15-puzzle, which involves actions such as \{(U), (D), (L), (R), (UL), (UR), (DL), (DR)\}, we generate domain variations by randomly selecting the set of actions available for each cell within the puzzle. We concatenate the domain's action space information with the state representation and input is fed into the heuristic function \( h(s, va) \), where \( va \) represents the action space information. 

This additional context allows heuristic function to more accurately predict cost-to-go values, even in previously unseen domains. Thus, we can present a new domain to our deep neural network with the action space information, facilitating development of heuristic function that are effective within a single domain and adaptable across various domains. This approach advances the field of pathfinding in AI, showcasing potential for domain generalizability through Deep RL.

\section{Methodology}

This section details techniques used to generate dynamic puzzle domains and updated approach to approximate value iteration. We also outline the evaluation metrics employed to assess correlation between learned and ground truth heuristic values.

\subsection{Environment Generator}

This section outlines the Algorithm \ref{alg:high-level-env-init} (Appendix \ref{subsec:envgen}) used to generate dynamic puzzle domains by initializing randomized actions for each cell and ensuring the availability of reversible actions. The environment generator assigns a random set of actions to each cell in the puzzle grid and validates their reversibility to ensure that the environment supports bidirectional navigation.

Reversible moves are crucial as they ensure every action in the environment has a counteraction to revert the state, confirming the puzzle's solvability from any configuration. We use reversible moves to obtain ground truth heuristic values, as having unsolvable states (infinities) would complicate determining these values. However, the proposed model can also solve domains without reversible moves; this approach is chosen for convenience in this study.

For the 8-puzzle, 8 actions across 9 cell positions yield 72 possible maps for defining available actions. Reversible moves fill 18 map slots (9 cells \(\times\) 2 actions), leaving \(72 - 18 = 54\) slots, resulting in \(2^{54}\) possible domain variations. Similarly, the 15-puzzle with 8 actions and 16 cell positions yields \(2^{96}\) configurations, and the 24-puzzle with 8 actions and 25 cell positions gives \(2^{150}\) configurations. This method ensures a comprehensive set of domains for effective model training and evaluation.

\subsection{Proposed Approximate Value Iteration}

In the updated approach to approximate value iteration, we enhance the state representation by concatenating it with a one-hot encoding of the action space. This combined representation provides additional context to the heuristic function about the available actions at each state, leading to more accurate predictions. The updated cost-to-go function now takes this augmented representation into account, as shown in the equation below:
\begin{equation}
    h'(s,va) = \min_a \left( c(s, T(s, a)) + h(T(s, a), va) \right)
    \label{eq:upavi}
\end{equation}
This modification allows the heuristic function to leverage state and action information, thereby improving the generalization and accuracy of the cost-to-go estimations.

\section{Evaluation Metrics}

The correlation between the learned heuristic value and the ground truth heuristic value is evaluated using two statistical measures: \textbf{1. Concordance Correlation Coefficient (CCC)}: CCC measures the agreement between two variables (learned and true heuristic values) by considering both precision and accuracy, capturing how closely the predictions match the ground truth. \textbf{2. Coefficient of Determination (R-squared, \( R^2 \))}: \( R^2 \) quantifies the proportion of variance in the ground truth heuristic values that is predictable from the learned heuristic values, providing insight into the goodness of fit of the model. More details in the Appendix \ref{subsec:evaluationmetrics}

\section{Experimental Setup}

Our experimental setup is designed to evaluate the effectiveness of incorporating action space information into training heuristic functions for solving n-puzzle problems. We conduct several experiments across domains within 8-puzzle, 15-puzzle and 24-puzzle, comparing models' performance with different inputs and training paradigms. Hardware details are in Appendix \ref{sec:hardware}.

\subsection{Model Architecture}

We use a Residual Network (ResNet) \citep{he2016deep} architecture implemented using PyTorch \citep{paszke2019pytorch}. The architecture includes an input layer for state or combined state-action space, initial processing layers that expand the input to a hidden space of size 5000 with ReLU activation, and four residual blocks, each operating in a 1000-dimensional hidden space, to facilitate deep learning without performance degradation. The output layer then transforms the processed data into heuristic estimates.

\subsection{Training Variations}



The experiments assessed the impact of different training setups by training the original DeepCubeA model on three n-puzzle domains: canonical actions (C) only (U, D, L, R), diagonal actions (D) only (UL, UR, DL, DR), and all actions (C+D) combined for 8-puzzle, 15-puzzle and 24-puzzle. DeepCubeA, operating on the original deep value iteration (Equation \ref{eq:orgavi}), observes 10 billion examples during training due to the constant action space, allowing for faster generation.

In contrast, the proposed model uses approximately 1 billion examples, with each state generated from a new action space variation domain, resulting in slower generation. This model was trained using approximate value iteration with state transition information (Equation \ref{eq:upavi}) for the 8-puzzle, 15-puzzle, and 24-puzzle. Additionally, we trained another model on each state example generated from a new action space variation domain using approximate value iteration without state transition information (Equation \ref{eq:orgavi}) for the 8-puzzle and 15-puzzle.

\subsection{Test Data Generation}
To assess the proposed model's performance, we generated three test datasets: \textbf{(Data1)} with 500 states per domain (C, D, and C+D) through random walks of 1000 to 10000 steps from the goal state; \textbf{(Data2)} with 100 states per domain across 500 unseen action space variation domains through random walks of 0 to 500 steps from the goal state; and \textbf{(Data3)} with 1500 states per domain (C, D, and C+D) through random walks of 0 to 500 steps from the goal state to compare the heuristic values against the optimal ground truth heuristic value. These data thoroughly compares the models' performance and adaptability across different action spaces.

Using the Fast Downward planner with the merge-and-shrink heuristic \citep{sievers2018merge}, we obtained the ground truth heuristic values for all generated test data. This heuristic simplifies the state space by merging and shrinking states, providing optimal heuristics. However, obtaining ground truth for complex puzzles like the 24-puzzle is computationally intensive. For example, \textbf{Data3} (1500 states), the process of obtaining ground truth heuristic value took over 3 weeks, running 15 problems in parallel across 100 instances, each problem with a 144-hour cutoff. This underscores the significant computational resources and time required for large-scale puzzles. More in Appendix \ref{sec:gtcompute}.

\section{Results}
\subsection{Comparison Against Ground Truth}
To quantitatively assess the performance of the trained heuristic model, we employ several metrics that reflect both the accuracy of the heuristic values and the efficiency of the pathfinding process.
\begin{figure}[!htbp]
    \centering
    \begin{subfigure}[b]{0.25\textwidth}
        \centering
        \includegraphics[width=\textwidth]{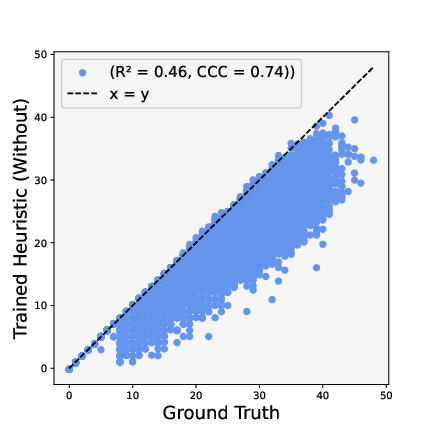}
        \caption{Without Action Info}
        \label{fig:15without}
    \end{subfigure}
    \begin{subfigure}[b]{0.25\textwidth}
        \centering
        \includegraphics[width=\textwidth]{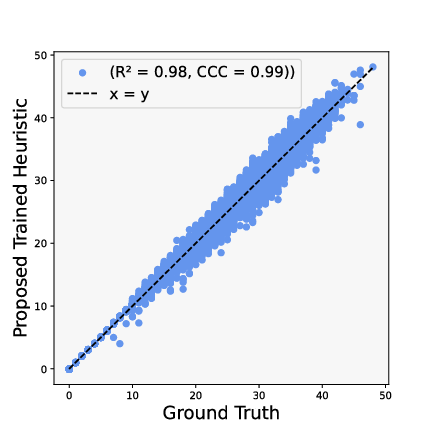}
        \caption{With Action Info}
        \label{fig:15with}
    \end{subfigure}
    \caption{Comparison of heuristic values predicted by the proposed model (\ref{fig:15with}) and the model without action information (\ref{fig:15without}) against ground truth heuristic values for 15-puzzle. The model with action information performs significantly better.}
    \label{fig:proavi_15}
\end{figure}
We generated \textbf{(Data2)} for the 15-puzzle domain, plotting the ground truth heuristic values against the predicted heuristic values for both the proposed model (with action information) and the model trained without action information, as shown in Figure \ref{fig:proavi_15}. The proposed model achieved a CCC of 0.99 and \( R^2 \) of 0.98, while the model without action information achieved a CCC of 0.74 and \( R^2 \)
\begin{figure}[!htbp]
    \centering
    \begin{subfigure}[b]{0.25\textwidth}
        \centering
        \includegraphics[width=\textwidth]{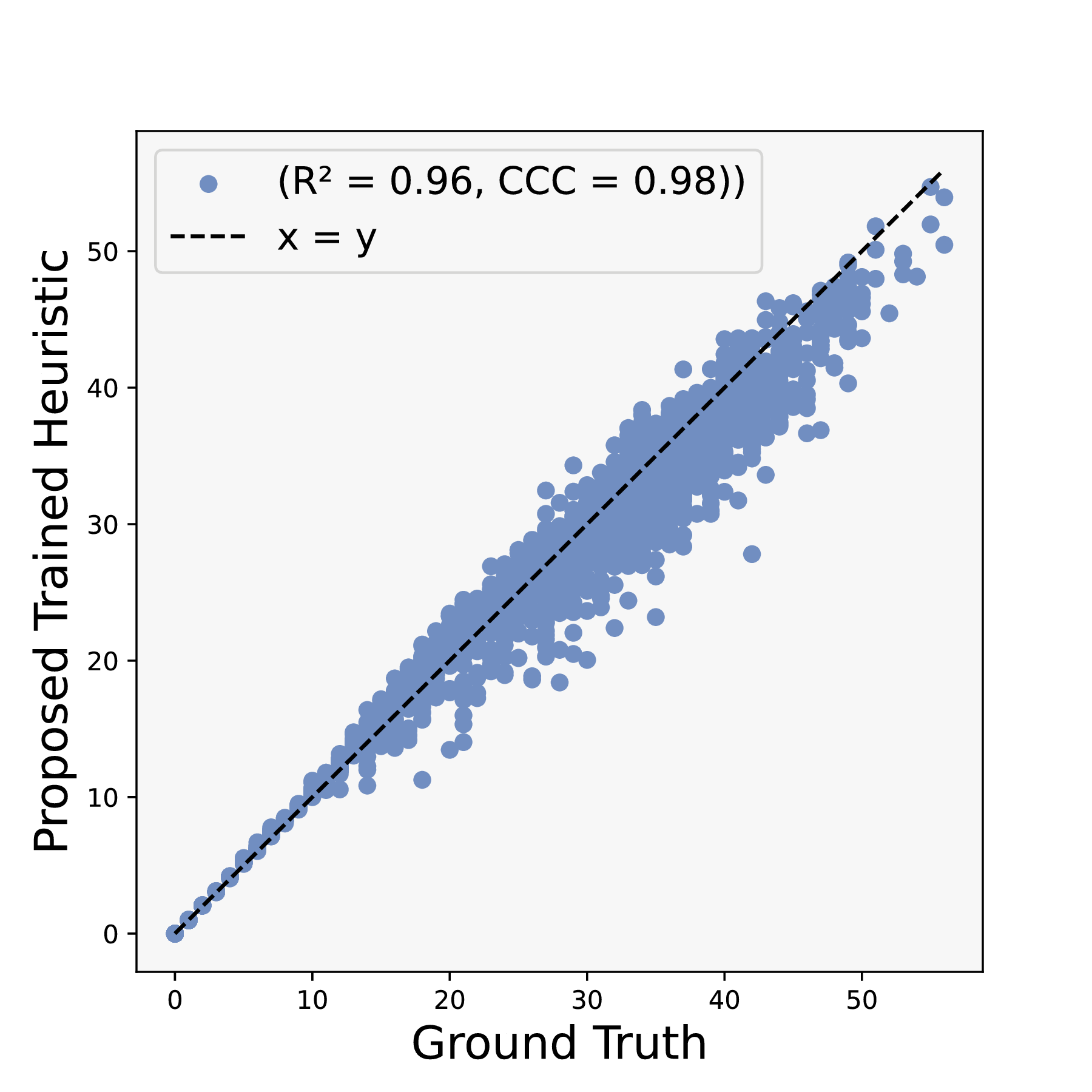} 
        \caption{C: P vs GT}
        \label{fig:comppvsgtanddcavsgt15puzsub1}
    \end{subfigure}
    \begin{subfigure}[b]{0.25\textwidth}
        \centering
        \includegraphics[width=\textwidth]{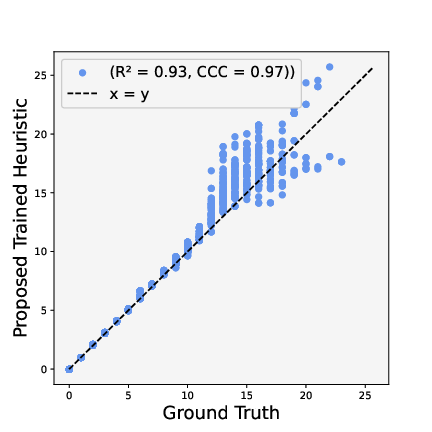} 
        \caption{D: P vs GT}
        \label{fig:comppvsgtanddcavsgt15puzsub2}
    \end{subfigure}
    \begin{subfigure}[b]{0.25\textwidth}
        \centering
        \includegraphics[width=\textwidth]{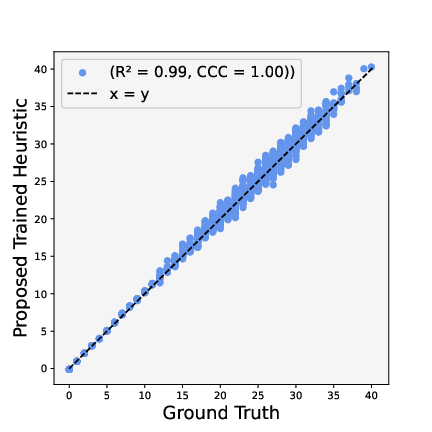} 
        \caption{C+D: P vs GT}
        \label{fig:comppvsgtanddcavsgt15puzsub3}
    \end{subfigure}

    \begin{subfigure}[b]{0.25\textwidth}
        \centering
        \includegraphics[width=\textwidth]{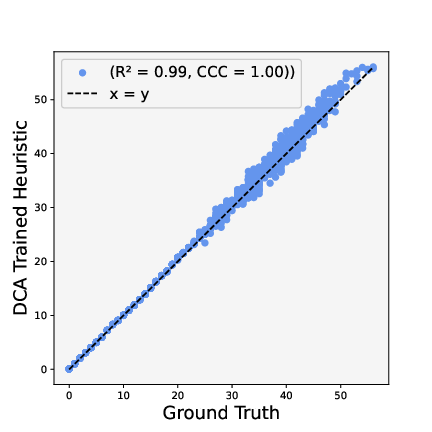} 
        \caption{C: DCA vs GT}
        \label{fig:comppvsgtanddcavsgt15puzsub4}
    \end{subfigure}
    \begin{subfigure}[b]{0.25\textwidth}
        \centering
        \includegraphics[width=\textwidth]{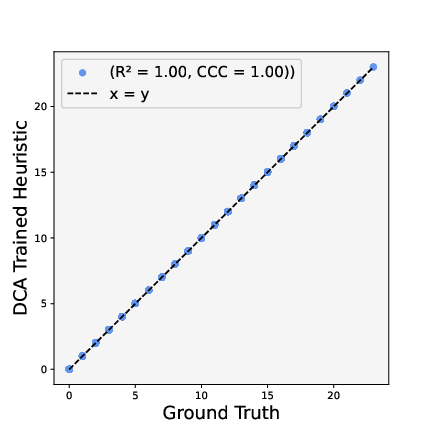} 
        \caption{D: DCA vs GT}
        \label{fig:comppvsgtanddcavsgt15puzsub5}
    \end{subfigure}
    \begin{subfigure}[b]{0.25\textwidth}
        \centering
        \includegraphics[width=\textwidth]{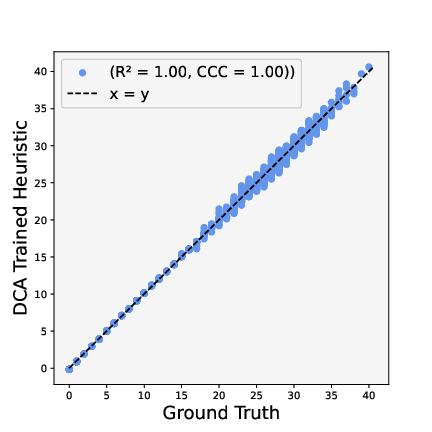} 
        \caption{C+D: DCA vs GT}
        \label{fig:comppvsgtanddcavsgt15puzsub6}
    \end{subfigure}  
    \caption{Comparison of trained and ground truth (GT) heuristic values for the 15-puzzle domain. \ref{fig:comppvsgtanddcavsgt15puzsub1}, \ref{fig:comppvsgtanddcavsgt15puzsub2}, \ref{fig:comppvsgtanddcavsgt15puzsub3} for the proposed model (P), and \ref{fig:comppvsgtanddcavsgt15puzsub4}, \ref{fig:comppvsgtanddcavsgt15puzsub5}, \ref{fig:comppvsgtanddcavsgt15puzsub6} for DeepCubeA (DCA) variants, each showing heuristic values for 1500 states across canonical actions (C), diagonal actions (D), and canonical + diagonal actions (C+D).}
    \label{fig:comppvsgtanddcavsgt15puz}
\end{figure}
 of  0.46, demonstrating the superior performance of the proposed model. In the 15-puzzle domain (Figure \ref{fig:comppvsgtanddcavsgt15puz}), DCA graphs show R² and CCC of 1.0 for all domains, except for the C only domain where R² is 0.99 and CCC is 1.0. The proposed model achieves R² of 0.96 and CCC of 0.98 for C, R² of 0.93 and CCC of 0.97 for D, and near-perfect correlation (R² of 0.99 and CCC of 1.0) for C+D. The proposed model shows robust performance across varying action domains. Results with 8-puzzle and 24-puzzle are in Appendix \ref{sec:resultsapend}.


\subsection{Performance Comparison}
To assess the proposed model's performance compared to domain-specific DeepCubeA variants and the Fast Downward Planner using the Fast Forward heuristic (FD(FF)), we evaluated across puzzle sizes (8-puzzle, 15-puzzle, and 24-puzzle) and action domains (C, D, and C+D) using performance metrics. We generated \textbf{Data1} for each domain variation. The Fast Forward heuristic with A* search had a 200-second time limit. For DeepCubeA and the proposed model, we performed a weighted A* search with a weight of 0.8 and a batch size of 1000, also with a 200-second time limit.

\begin{table}[h]
    \begin{center}
        \begin{tabular}{|p{2.8cm}|p{1.7cm}|p{0.8cm}|p{1cm}|p{1.6cm}|p{1cm}|p{1.72cm}|p{0.96cm}|}
            \hline
            \textbf{Domain} & \textbf{Solver} & \textbf{Len} & \textbf{Opt} & \textbf{Nodes} & \textbf{Secs} & \textbf{Nodes/Sec} & \textbf{Solved} \\
            \hline
            8 Puzzle \textbf{(C)} & DeepCubeA & \textbf{18.38} & \textbf{100\%} & 3.59E+04 & 0.69  & \textbf{5.2E+04} & \textbf{100\%} \\ 
            8 Puzzle \textbf{(C)} & Proposed & \textbf{18.38} & \textbf{100\%} & 7.17E+04 & 1.76  & 4.07E+04 &\textbf{100\%} \\ 
            8 Puzzle \textbf{(C)} & FD(FF) & 18.8 & 81\% & \textbf{5.56E+02} & \textbf{0.11} & 4.7E+03 & \textbf{100\%} \\ 
            \hline
            8 Puzzle \textbf{(D)} & DeepCubeA & \textbf{1.44} & \textbf{100\%} & 1.95E+01 & \textbf{0.01} & 2.92E+03 & \textbf{100\%} \\ 
            8 Puzzle \textbf{(D)} & Proposed & \textbf{1.44} & \textbf{100\%} & 4.05E+01 & \textbf{0.01} & \textbf{4.92E+03} & \textbf{100\%} \\ 
            8 Puzzle \textbf{(D)} & FD(FF) & \textbf{1.44} & \textbf{100\%} & \textbf{2.45E+00} & 0.2 & 1.23E+01 & \textbf{100\%} \\ 
            \hline
            8 Puzzle \textbf{(C+D)} & DeepCubeA & \textbf{11.84} & \textbf{100\%} & 6.2E+04 & \textbf{1.18} & \textbf{5.26E+04} & \textbf{100\%} \\ 
            8 Puzzle \textbf{(C+D)} & Proposed & \textbf{11.84} & \textbf{100\%} & 6.23E+04 & 1.56 & 3.97E+04 & \textbf{100\%}  \\ 
            8 Puzzle \textbf{(C+D)} & FD(FF) & 12.9 & 54.2\% & \textbf{8.68E+01} & \textbf{0.13} & 6.59E+02 & \textbf{100\%} \\ 
            \hline
            \hline
             15 Puzzle \textbf{(C)} & DeepCubeA & \textbf{52.03} & \textbf{99.4\%}& \textbf{1.82E+05 }& \textbf{4.31} & 4.21E+04 & \textbf{100\%} \\ 
            15 Puzzle \textbf{(C)} & Proposed & 52.18 & 93.76\% & 3.62E+05 & 10.39 & 3.49E+04 & \textbf{100\%} \\ 
            15 Puzzle \textbf{(C)} & FD(FF) & 52.75 & 24.80 & 2.92E+06 & 42.11 & \textbf{6.93E+04} & 80.80\% \\ 
            \hline
            15 Puzzle \textbf{(D)} & DeepCubeA & \textbf{10.8} & \textbf{100\%} & 8.2E+02 & \textbf{0.03} & 2.43E+04 & \textbf{100\%} \\ 
            15 Puzzle \textbf{(D)} & Proposed & 10.81 & 99.8\% & 1.64E+03 & 0.05 & \textbf{3.01E+04} & \textbf{100\%} \\ 
            15 Puzzle \textbf{(D)} & FD(FF) & 10.86 & 96.8\% & \textbf{4.18E+01} & 0.21 & 1.96E+02 & \textbf{100\%} \\ 
            \hline
            15 Puzzle \textbf{(C+D)} & DeepCubeA & \textbf{25.66} & \textbf{100\%} & 1.78E+05 & 3.74 & \textbf{4.78E+04}& \textbf{100\%} \\ 
            15 Puzzle \textbf{(C+D)} & Proposed & 25.67 & 99.8\% & 1.78E+05 & 4.72 & 3.78E+04 & \textbf{100\%} \\ 
            15 Puzzle \textbf{(C+D)} & FD(FF) & 29.32 & 13.4\% & \textbf{8.4E+03} & \textbf{1.17} & 3.56E+03 & \textbf{100\%} \\ 
            \hline
            \hline
             24 Puzzle \textbf{(C)} & DeepCubeA & \textbf{89.48} & \textbf{96.98\%} & \textbf{3.34E+05}& \textbf{8.05} & \textbf{4.15E+04} & \textbf{100\%} \\ 
            24 Puzzle \textbf{(C)} & Proposed & 92.80 & 22.03\% & 7.6E+05& 24.06 & 3.16E+04 & \textbf{100\%} \\
            24 Puzzle \textbf{(C)} & FD(FF) & 81.00 & 0.40 & 2.68E+06 & 89.84 & 2.91E+04 & 1.01\%\\
            \hline
            24 Puzzle \textbf{(D)} & DeepCubeA & \textbf{14.9} & \textbf{100\%} & 2.55E+04& \textbf{0.47} & \textbf{5.46E+04} & \textbf{100\%} \\ 
            24 Puzzle \textbf{(D)} & Proposed & 14.92 & 99.8\% & 5.1E+04& 1.35 & 3.78E+04 & \textbf{100\%} \\ 
            24 Puzzle \textbf{(D)} & FD(FF) & 15.16 & 89.2\% & \textbf{2.64E+02} & \textbf{0.12} & 2.05E+03 & \textbf{100\%} \\
            \hline
            24 Puzzle \textbf{(C+D)} & DeepCubeA & \textbf{31.33} & \textbf{100\%} & 2.27E+05& \textbf{4.83} & \textbf{4.69E+04} & \textbf{100\%} \\ 
            24 Puzzle \textbf{(C+D)} & Proposed & 31.34 & 99.6\% & 2.27E+05& 6.78 & 3.34E+04 & \textbf{100\%} \\ 
            24 Puzzle \textbf{(C+D)} & FD(FF) & 36.81 & 13.8\% & \textbf{1.7E+04} & 5.35 & 1.77E+03 & 99.4\%\\
            \hline
        \end{tabular}
    \end{center}
    \caption{Performance comparison of DeepCubeA action variants, the proposed model, and the Fast Downward Planner using the Fast Forward heuristic (FD(FF)). Metrics include average solution length (Len), optimality (Opt), average nodes generated (Nodes), average computation time (Secs), nodes processed per second (Nodes/Sec), and percentage of problems solved (Solved). Action variants cover (C), (D), and (C+D) actions for the 8-puzzle, 15-puzzle, and 24-puzzle domains. The proposed model performs comparably to DeepCubeA, and better than Fast Downward Planner.}

    \label{tab:comparison}
\end{table}



Table \ref{tab:comparison} highlights that the domain-independent fast-forward heuristic significantly underperforms compared to the proposed model and DeepCubeA. For instance, in the 15-puzzle (Canonical) domain, the FD(FF) solver solved only 80.8\% of the problems, whereas both the proposed model and DeepCubeA achieved 100\% success rates. Even when FD(FF) solves 100\% of the problems, its optimality is less than the learned heuristic functions.

The DeepCubeA variants were trained separately for each canonical (C), diagonal (D), and combined (C+D) domain for the 8-puzzle, 15-puzzle, and 24-puzzle. In contrast, the proposed model was trained only once per puzzle size, yet it exhibits competitive performance. For example, in the 8-puzzle (Canonical) domain, the proposed model's average solution length was identical to DeepCubeA. In the 24-puzzle (Canonical) domain, the proposed model's average solution length was only about 3.7\% longer than DeepCubeA's, showcasing its efficiency and generalization capability despite using only 10\% of the training data compared to DeepCubeA.

\section{Discussion and Future Work}



To the best of our knowledge, this is the first study using reinforcement learning to achieve generalization across pathfinding domains. Leveraging Deep Approximate Value Iteration (DAVI), our approach integrates state transition information into state representations, enabling models to generalize across various 15-puzzle action space variation domains, a capability not present in previous works like DeepCubeA. This method allows the heuristic function to predict cost-to-go values for unseen domains without retraining or fine-tuning. The proposed model's strong performance and competitive metrics against DeepCubeA variants highlight the potential of combining state and state transition information in reinforcement learning.

Future work will explore incorporating advanced techniques such as \citep{chen2023goose}, which employs graph neural networks on abstract graph representations of problems using PDDL. While PDDL requires an explicit definition of the transition function, encoding state-action-next state tuples into embeddings for the heuristic function presents a powerful alternative. Additionally, leveraging knowledge graphs to encode transitions could significantly enhance heuristic functions. This approach is especially potent when knowledge graphs can be automatically derived or manually adjusted. For instance, in robotics, a human operator could tweak the knowledge graph when a robot behaves unexpectedly, allowing the heuristic function to adapt seamlessly. These methods promise to improve the generalization capabilities and robustness of heuristic functions in solving complex pathfinding problems across various domains.

\section{Conclusion}

In this preliminary work, we have integrated state transition information with state representations to train heuristic functions capable of generalizing across different 15-puzzle action space variation domains. Our approach demonstrates that incorporating state transition information enables the heuristic function to predict cost-to-go values in previously unseen domains without fine-tuning. Comparative analysis with DeepCubeA variants shows that our model performs competitively despite training on a smaller dataset. These findings highlight the potential of our method to enhance generalization and efficiency in pathfinding problems. Future work will focus on incorporating transition dynamics to improve the model's domain understanding, exploring generalization across various puzzle domains, and employing advanced techniques such as graph neural networks and knowledge graphs to further enhance the adaptability and robustness of heuristic functions.

\section*{Broader Impact Statement}
\label{sec:broaderImpact}
The broader impact of this research highlights the significant computational resources required to train deep neural networks for pathfinding problems. By developing models that generalize across various domains, we can reduce the need for repeated, resource-intensive training processes, thus saving substantial computational power and minimizing environmental impact. This approach promotes efficiency and sustainability in developing AI solutions for pathfinding and beyond.



\bibliography{mybib,main}
\bibliographystyle{rlc}

\appendix

\section{Background and Literature Review}

\subsection{N-Puzzle}\label{subsec:npuzz}

\begin{figure}[!htbp]
    \centering
    \begin{subfigure}[b]{0.2\textwidth}
        \centering
        \includegraphics[width=\textwidth]{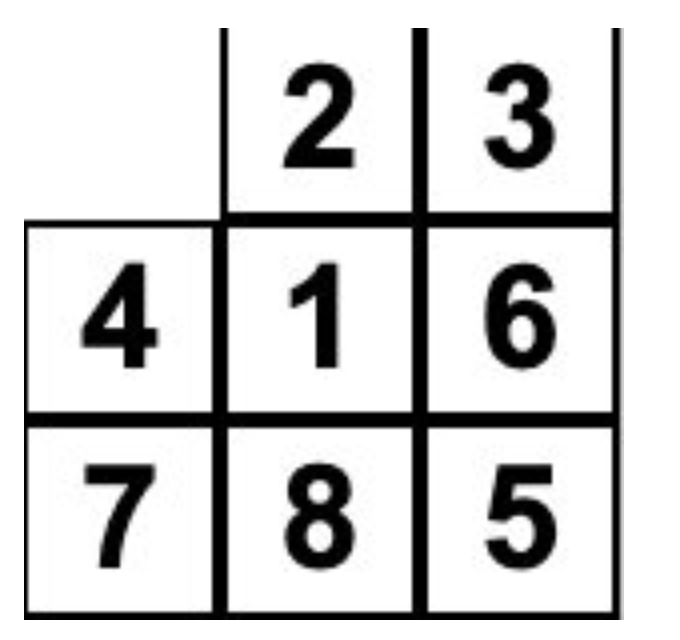}
        \caption{Scrambled State}
        \label{fig:npuz_scrambled}
    \end{subfigure}
    \begin{subfigure}[b]{0.2\textwidth}
        \centering
        \includegraphics[width=\textwidth]{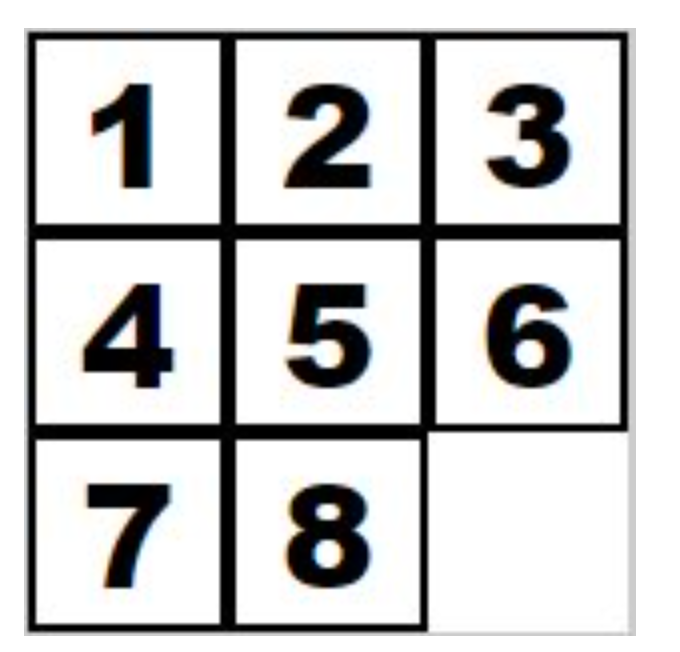}
        \caption{Goal State}
        \label{fig:npuz_goal}
    \end{subfigure}
   \caption{Example of a scrambled state and the goal state for the 8-puzzle domain. The cost-to-go is significantly reduced when including diagonal moves: 16 steps for canonical moves only versus 2 steps for canonical and diagonal moves combined.}
    \label{fig:npuz}
\end{figure}

The n-puzzle, also known as the sliding tile puzzle \citep{keith2011vintage}, is a classic problem in artificial intelligence. The puzzle consists of a grid of \(n + 1 \) tiles, where \( n \) tiles are numbered sequentially, and one tile is left empty. The objective is to rearrange the tiles from a given initial state to a goal state, typically with tiles ordered sequentially from top-left to bottom-right, and the empty tile in the bottom-right corner, as shown in Figure \ref{fig:npuz_goal}. Traditionally, actions correspond to moving a tile into the empty space, which can be one of the following moves: up (U), down (D), left (L), and right (R). These moves define the canonical action set. In addition to the canonical moves, we consider diagonal actions: upper-left (UL), upper-right (UR), lower-left (DL), and lower-right (DR). These additional moves increase the possible transitions from any given state. Figure \ref{fig:npuz} illustrates an example of a scrambled state and the goal state.

\subsection{Large Language Models in Pathfinding Problems}\label{subsec:llms}

Large language models (LLMs) are a foundation model pre-trained on extensive textual datasets using self-supervised learning techniques. These models, characterized by their ability to generate and comprehend human-like text, have shown potential in various natural language processing (NLP) tasks. LLMs can be fine-tuned as foundation models for several downstream tasks, such as machine translation and question-answering. There have been several attempts to use LLMs for solving pathfinding problems and generating a sequence of actions to solve the problem.

 While attempts to use causal LLMs to solve pathfinding problems via
in-context learning have shown modest performance and indicates notable challenges \cite{sermanet2023robovqa, li2023human, silver2023generalized, parakh2023human, zelikman2023parsel, besta2023graph, huang2023voxposer, dalal2023plan, wang2023plan, valmeekam2022large, valmeekam2023can, gramopadhye2022generating, singh2023progprompt}. Other prompting techniques have been introduced to enhance LLMs’ reasoning capabilities \cite{hu2023chain, yao2023tree}.

Efforts to generate multimodal, text, and image-based sequences of actions are exemplified by \cite{lu2023multimodal}. Additionally, a subset of studies investigates the fine-tuning of seq2seq, code-based language models \cite{pallagani2022plansformer, pallagani2023plansformer}, which are noted for their advanced syntactic encoding. These models show promise within the confines of their training datasets \cite{logeswaran2023code} yet exhibit limitations in generalizing to out-of-distribution domains \cite{pallagani2023understanding}, highlighting a gap in their adaptability across diverse pathfinding problem domains.

One of the major drawbacks of LLMs is that they do not inherently possess search capabilities, nor do these papers employ heuristic search techniques. This limitation presents a significant challenge in applying LLMs to pathfinding problems, as effective search strategies are crucial for optimal solutions.

\subsection{Domain-Independent Planning}\label{subsec:domainindeependent}

The Fast Downward planner \cite{helmert2006fast} is a well-known automated planning system that utilizes various heuristics to solve complex pathfinding problems. One key heuristic employed by Fast Downward is the Fast Forward (FF) heuristic \cite{hoffmann2001ff}. The FF heuristic estimates the cost-to-go by ignoring the delete effects of actions, allowing it to compute heuristic values with minimal computational overhead rapidly. These heuristics are domain-independent, meaning they can generalize across various planning domains without requiring domain-specific adjustments. This generalizability makes them highly versatile and efficient for various applications. However, despite their efficiency, domain-independent heuristics do not perform as well as learned heuristics \cite{agostinelli2024specifying}. This underscores the limitations of domain-independent heuristics and highlights the potential benefits of leveraging training to achieve superior pathfinding performance.

\section{Methodology}
\label{sec:algenv}

\subsection{Environment Generator} \label{subsec:envgen}

This section lists the environment generator algorithm.

\begin{algorithm}
\caption{Environment Generator - NPuzzle with action variation}
\label{alg:high-level-env-init}
\begin{algorithmic}[1]
\State \textbf{Input}: $n$ - dimension of the puzzle grid (e.g., $n=3$ for an 8-puzzle)
\State \textbf{Output}: Puzzle environment $P$ with reversible actions validated

\Procedure{InitializeAndValidatePuzzle}{$n$}
    \State Let $P$ be an $n \times n$ grid of cells
    \State Define $A = \{\text{U, D, L, R, UL, UR, DL, DR}\}$ as the set of all possible actions
    \For{each cell $c_{ij} \in P$}
        \State Assign to $c_{ij}.A$ a random subset of $A$
    \EndFor
    \For{each cell $c_{ij} \in P$}
        \For{each action $a \in c_{ij}.A$}
            \State Determine the target cell $c_{kl}$ from $c_{ij}$ using action $a$
            \State Identify the reverse action $a'$ corresponding to $a$
            \If{$a' \notin c_{kl}.A$}
                \State Remove $a$ from $c_{ij}.A$
            \EndIf
        \EndFor
    \EndFor
    \State \Return $P$
\EndProcedure
\end{algorithmic}
\end{algorithm}

\subsection{Evaluation Metrics} \label{subsec:evaluationmetrics} 
The correlation between the learned heuristic value and the ground truth heuristic value is evaluated using two statistical measures:\\
    \noindent \textbf{1. Concordance Correlation Coefficient (CCC)}: Denoted as \( \rho_c \), CCC measures the agreement between two variables (learned and true heuristic values), considering both the precision and accuracy \cite{lawrence1989concordance}. It is defined as:
    \[
    \rho_c = \frac{2 \rho \sigma_x \sigma_y}{\sigma_x^2 + \sigma_y^2 + (\mu_x - \mu_y)^2}
    \]
    Where \( \rho \) is the Pearson correlation coefficient between the variables, \( \sigma_x \) and \( \sigma_y \) are the standard deviations, and \( \mu_x \) and \( \mu_y \) are the means of the true and predicted values, respectively.\\ This metric quantifies the proportion of variance in the dependent variable (ground truth heuristic values) that is predictable from the independent variable (predicted heuristic values). It primarily measures the trend of the predictive accuracy but does not account for how close the predictions are to the identity line (perfect agreement).
    
    \noindent \textbf{2. Coefficient of Determination (R-squared, \( R^2 \))}: This metric quantifies the proportion of variance in the dependent variable that is predictable from the independent variable, providing insight into the goodness of fit of the model to the data \cite{chicco2021coefficient}. 
    \[
    R^2 = 1 - \frac{\sum_{i} (y_i - \hat{y}_i)^2}{\sum_{i} (y_i - \bar{y})^2}
    \]
    where \( y_i \) are the ground truth values, \( \hat{y}_i \) are the predicted heuristic values, and \( \bar{y} \) is the mean of the ground truth values.\\ This metric assesses the agreement between two variables, considering the precision and the accuracy. This metric captures the relationship trend between predicted and true values and evaluates how closely the predictions conform to the identity line, indicating perfect prediction accuracy.

\section{Experimental Setup}

\subsection{Hardware Details} \label{sec:hardware}
We have used two servers to run our experiments: one with 48-core nodes each hosting 2 V100 32G GPUs and 128GB of RAM, and another with 256 cores, eight A100 40GB GPUs, and 1TB of RAM. The processor speed is 2.8 GHz.

\subsection{Test Data Generation}\label{sec:gtcompute}

To assess the proposed model's performance, we generated three test datasets:
\begin{itemize}
    \item \textbf{Data1}: 500 states per domain (Canonical (C), Diagonal (D), and Canonical+Diagonal (C+D)) for 8-puzzle, 15-puzzle, and 24-puzzle, created through random walks of 1000 to 10000 steps from the goal state.
    \item \textbf{Data2}: 100 states per domain across 500 unseen action space variation domains for 8-puzzle and 15-puzzle, created through random walks of 0 to 500 steps from the goal state.
    \item \textbf{Data3}: 1500 states per domain (C, D, and C+D) for 8-puzzle, 15-puzzle, and 24-puzzle, created through random walks of 0 to 500 steps from the goal state to compare heuristic values against the optimal ground truth heuristic value.
\end{itemize}

These datasets were essential for thoroughly comparing the models’ performance and adaptability across different action spaces.

Using the Fast Downward planner with the merge-and-shrink heuristic \citep{sievers2018merge}, we obtained the ground truth heuristic values for all generated test data. This heuristic simplifies the state space by merging and shrinking states, providing optimal heuristics. However, obtaining ground truth for complex puzzles like the 24-puzzle is computationally intensive.

For \textbf{Data1}:
\begin{itemize}
    \item \textbf{8-puzzle}: The process took approximately one day with a cutoff of 160 minutes per problem, running one instance of the planner per domain.
    \item \textbf{15-puzzle}: The process took about a week with a cutoff of 160 minutes per problem, running five instances of the planner per domain.
    \item \textbf{24-puzzle}: The process took around 2 weeks with a cutoff of 144 hours per problem, running 50 instances of the planner per domain.
\end{itemize}

For \textbf{Data2}:
\begin{itemize}
    \item \textbf{8-puzzle}: The process took one week with a cutoff of 160 minutes per problem, running 50 instances overall.
    \item \textbf{15-puzzle}: The process also took one week with a cutoff of 160 minutes per problem, running 500 instances overall.
\end{itemize}

For \textbf{Data3}:
\begin{itemize}
    \item \textbf{8-puzzle}: The process took approximately 2 days with a cutoff of 160 minutes per problem, running one instance of the planner per domain.
    \item \textbf{15-puzzle}: The process took about a week with a cutoff of 160 minutes per problem, running 15 instances of the planner per domain.
    \item \textbf{24-puzzle}: The process took over 3 weeks, running 100 instances per domain, with 15 problems per instance and a cutoff of 144 hours per problem.
\end{itemize}

These computations highlight the significant computational resources and time required to obtain ground truth heuristic values for large-scale puzzles, underscoring the challenges in benchmarking heuristic functions for complex domains.

\section{Results}
\label{sec:resultsapend}
To quantitatively assess the performance of the trained heuristic model, we employ several metrics that reflect both the accuracy of the heuristic values and the efficiency of the pathfinding process.

\begin{figure}[htbp]
    \centering
    \begin{subfigure}[b]{0.25\textwidth}
        \centering
        \includegraphics[width=\textwidth]{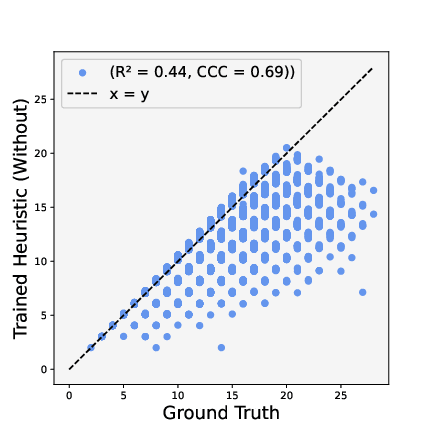}
        \caption{Without Action Info}
        \label{fig:8without}
    \end{subfigure}
    \begin{subfigure}[b]{0.25\textwidth}
        \centering
        \includegraphics[width=\textwidth]{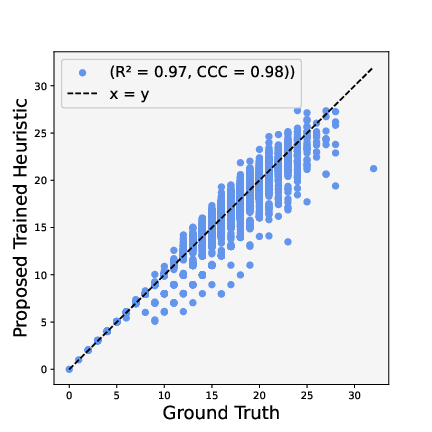}
        \caption{With Action Info}
        \label{fig:8with}
    \end{subfigure}
    \caption{Comparison of heuristic values predicted by the proposed model (\ref{fig:8with}) and the model without action information 
    (\ref{fig:8without}) against ground truth heuristic values for 8-puzzle. The model with action information performs significantly better.}
    \label{fig:proavi}
\end{figure}

We generate \textbf{(D2)} for the 8-puzzle domain. We plot the ground truth optimal heuristic values against the trained heuristic values for both the proposed model (with action information) and the model trained without action information, as shown in Figure \ref{fig:proavi}. The proposed model achieved a CCC of 0.98 and $ R2$ of 0.97, while the model without action information achieved a CCC of 0.69 and $ R2$ of 0.44, demonstrating the proposed model's superior performance.
\clearpage

\begin{figure}[htbp]
    \centering
    \begin{subfigure}[b]{0.25\textwidth}
        \centering
        \includegraphics[width=\textwidth]{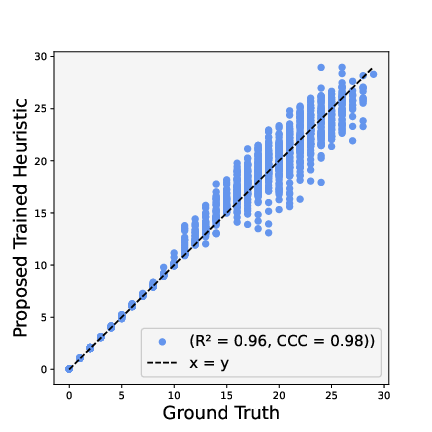}
        \caption{C: P vs GT}
        \label{fig:comppvsgtanddcavsgt8puzsub1}
    \end{subfigure}
    \begin{subfigure}[b]{0.25\textwidth}
        \centering
        \includegraphics[width=\textwidth]{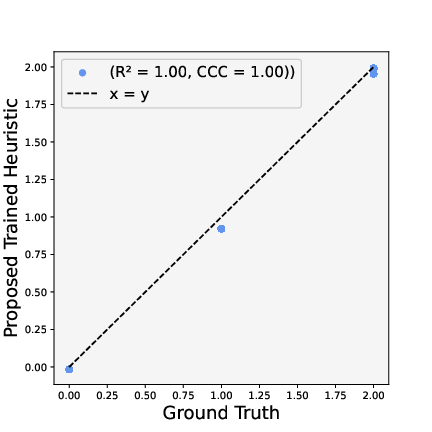}
        \caption{D: P vs GT}
        \label{fig:comppvsgtanddcavsgt8puzsub2}
    \end{subfigure}
    \begin{subfigure}[b]{0.25\textwidth}
        \centering
        \includegraphics[width=\textwidth]{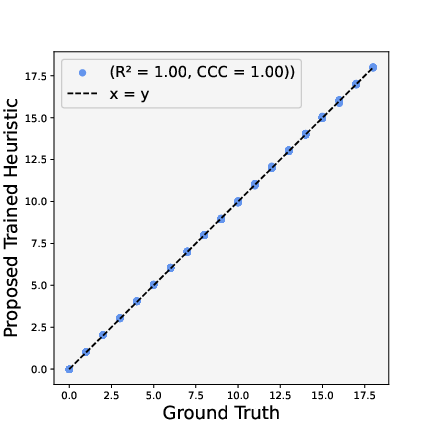} 
        \caption{C+D: P vs GT}
        \label{fig:comppvsgtanddcavsgt8puzsub3}
    \end{subfigure}
    
    \begin{subfigure}[b]{0.25\textwidth}
        \centering
        \includegraphics[width=\textwidth]{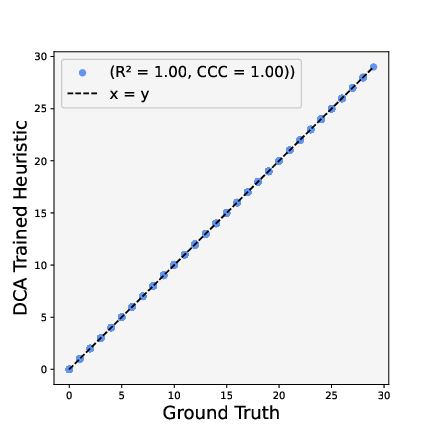}
        \caption{C: DCA vs GT}
        \label{fig:comppvsgtanddcavsgt8puzsub4}
    \end{subfigure}
    \begin{subfigure}[b]{0.25\textwidth}
        \centering
        \includegraphics[width=\textwidth]{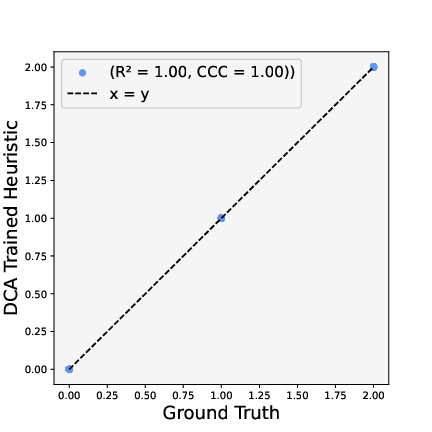}
        \caption{D: DCA vs GT}
        \label{fig:comppvsgtanddcavsgt8puzsub5}
    \end{subfigure}
    \begin{subfigure}[b]{0.25\textwidth}
        \centering
        \includegraphics[width=\textwidth]{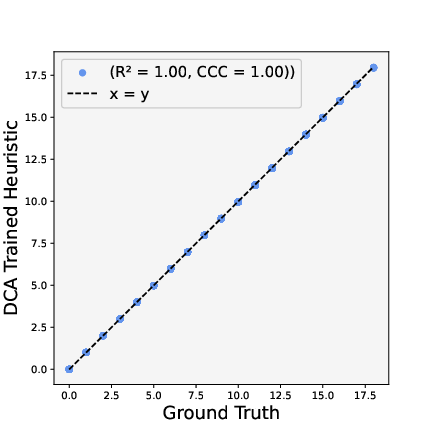} 
        \caption{C+D: DCA vs GT}
        \label{fig:comppvsgtanddcavsgt8puzsub6}
    \end{subfigure}
    
    \caption{Comparison of trained and ground truth (GT) heuristic values for the 8-puzzle domain. \ref{fig:comppvsgtanddcavsgt8puzsub1}, \ref{fig:comppvsgtanddcavsgt8puzsub2}, \ref{fig:comppvsgtanddcavsgt8puzsub3} for the proposed model (P), and \ref{fig:comppvsgtanddcavsgt8puzsub4}, \ref{fig:comppvsgtanddcavsgt8puzsub5}, \ref{fig:comppvsgtanddcavsgt8puzsub6} for DeepCubeA (DCA) variants, each showing heuristic values for 1500 states across canonical actions (C), diagonal actions (D), and canonical + diagonal actions (C+D).}
    \label{fig:comppvsgtanddcavsgt8puz}
\end{figure}

For the 8-puzzle domain (Figure \ref{fig:comppvsgtanddcavsgt8puz}), DCA graphs show a perfect correlation with R² and CCC of 1.0. The proposed model achieves an R² of 0.96 and CCC of 0.98 for the C only domain, and perfect correlation (R² and CCC of 1.0) for the D and C+D domains. The proposed model demonstrates near-perfect accuracy, especially in the more complex action domains.
\clearpage

\begin{figure}[htbp]
    \centering
    \begin{subfigure}[b]{0.25\textwidth}
        \centering
        \includegraphics[width=\textwidth]{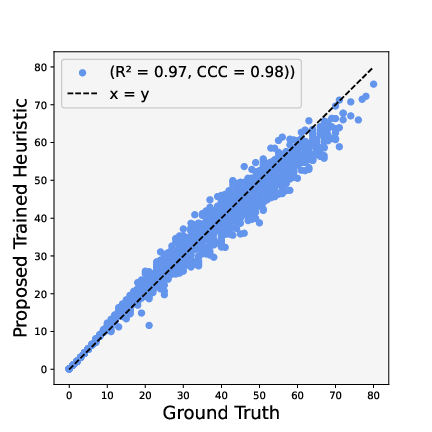} 
        \caption{C: P vs GT}
        \label{fig:comppvsgtanddcavsgt24puzsub1}
    \end{subfigure}
    \begin{subfigure}[b]{0.25\textwidth}
        \centering
        \includegraphics[width=\textwidth]{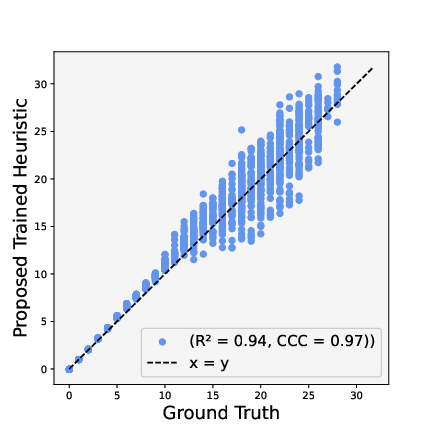} 
        \caption{D: P vs GT}
        \label{fig:comppvsgtanddcavsgt24puzsub2}
    \end{subfigure}
    \begin{subfigure}[b]{0.25\textwidth}
        \centering
        \includegraphics[width=\textwidth]{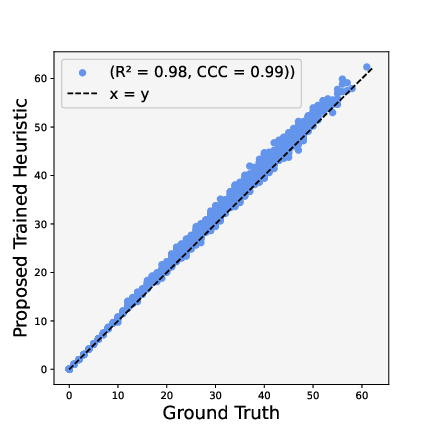} 
        \caption{C+D: P vs GT}
        \label{fig:comppvsgtanddcavsgt24puzsub3}
    \end{subfigure}
    
    \begin{subfigure}[b]{0.25\textwidth}
        \centering
        \includegraphics[width=\textwidth]{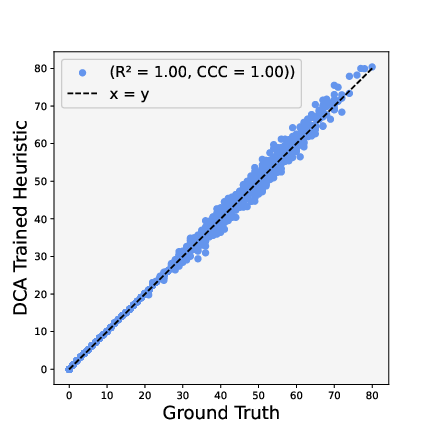} 
        \caption{C: DCA vs GT}
        \label{fig:comppvsgtanddcavsgt24puzsub4}
    \end{subfigure}
    \begin{subfigure}[b]{0.25\textwidth}
        \centering
        \includegraphics[width=\textwidth]{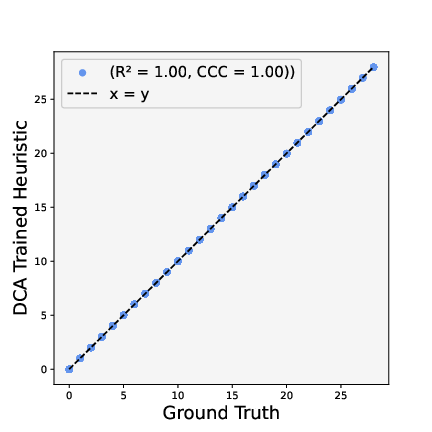} 
        \caption{D: DCA vs GT}
        \label{fig:comppvsgtanddcavsgt24puzsub5}
    \end{subfigure}
    \begin{subfigure}[b]{0.25\textwidth}
        \centering
        \includegraphics[width=\textwidth]{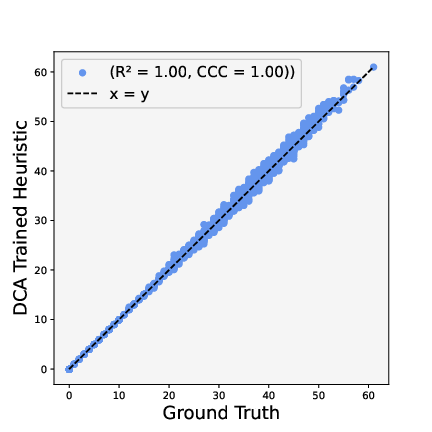} 
        \caption{C+D: DCA vs GT}
        \label{fig:comppvsgtanddcavsgt24puzsub6}
    \end{subfigure}
    
    \caption{Comparison of trained and ground truth (GT) heuristic values for the 24-puzzle domain. \ref{fig:comppvsgtanddcavsgt24puzsub1}, \ref{fig:comppvsgtanddcavsgt24puzsub2}, \ref{fig:comppvsgtanddcavsgt24puzsub3} for the proposed model (P), and \ref{fig:comppvsgtanddcavsgt24puzsub4}, \ref{fig:comppvsgtanddcavsgt24puzsub5}, \ref{fig:comppvsgtanddcavsgt24puzsub6} for DeepCubeA (DCA) variants, each showing heuristic values for 1500 states across canonical actions (C), diagonal actions (D), and canonical + diagonal actions (C+D).}
    \label{fig:comppvsgtanddcavsgt24puz}
\end{figure}

For the 24-puzzle domain (Figure \ref{fig:comppvsgtanddcavsgt24puz}), DCA graphs show perfect correlation (R² and CCC of 1.0). The proposed model achieves R² of 0.97 and CCC of 0.98 for C only, R² of 0.94 and CCC of 0.97 for D only, and near-perfect correlation (R² of 0.98 and CCC of 0.99) for C+D. The proposed model maintains high accuracy across all domains, demonstrating its generalizability.

\end{document}